\documentclass[10pt,twocolumn,letterpaper]{article}

\usepackage{cvpr}      

\definecolor{cvprblue}{rgb}{0.21,0.49,0.74}
\usepackage[pagebackref,breaklinks,colorlinks,allcolors=cvprblue]{hyperref}

\def\paperID{*****} 
\def\confName{CVPR}
\def\confYear{2025}

\title{NTIRE 2025 Challenge on UGC Video Enhancement: Methods and Results}

\author{Nikolay Safonov~\thanks{N.~Safonov (nikolay.safonov@graphics.cs.msu.ru), A.~Bryncev (alxbrc0@gmail.com), A.~Moskalenko (and.v.moskalenko@gmail.com), D.~Kulikov (dkulikov@graphics.cs.msu.ru), D.~Vatolin (dmitriy@graph ics.cs.msu.ru), and R.~Timofte (radu.timofte@uni-wuerzburg.de) were the challenge organizers, while the other authors participated in the challenge. \cref{affilations} contains the authors’ teams and affiliations. NTIRE 2025 webpage: \url{https://cvlai.net/ntire/2025/}} \and Alexey Bryncev \and Andrey Moskalenko \and Dmitry Kulikov \and Dmitry Vatolin \and Radu Timofte \and Haibo Lei \and Qifan Gao \and Qing Luo \and Yaqing Li \and Jie Song \and Shaozhe Hao \and Meisong Zheng \and Jingyi Xu \and Chengbin Wu \and Jiahui Liu \and Ying Chen \and Xin Deng \and Mai Xu \and Peipei Liang \and Jie Ma \and Junjie Jin \and Yingxue Pang \and Fangzhou Luo \and Kai Chen \and Shijie Zhao \and Mingyang Wu \and Renjie Li \and Yushen Zuo \and Shengyun Zhong \and Zhengzhong Tu}

\begin{document}
\maketitle
\begin{abstract}
This paper presents an overview of the NTIRE 2025 Challenge on UGC Video Enhancement.  The challenge constructed a set of 150 user-generated content videos without reference ground truth, which suffer from real-world degradations such as noise, blur, faded colors, compression artifacts, etc. The goal of the participants was to develop an algorithm capable of improving the visual quality of such videos. Given the widespread use of UGC on short-form video platforms, this task holds substantial practical importance.  The evaluation was based on subjective quality assessment in crowdsourcing, obtaining votes from over 8000 assessors. The challenge attracted more than 25 teams submitting solutions, 7 of which passed the final phase with source code verification. The outcomes may provide insights into the state-of-the-art in UGC video enhancement and highlight emerging trends and effective strategies in this evolving research area. All data, including the processed videos and subjective comparison votes and scores, is made publicly available — \url{https://github.com/msu-video-group/NTIRE25_UGC_Video_Enhancement}.



\end{abstract}    
\section{Introduction}
\label{sec:intro}

\begin{figure}[ht]
    \centering
    \includegraphics[width=1.0\columnwidth]{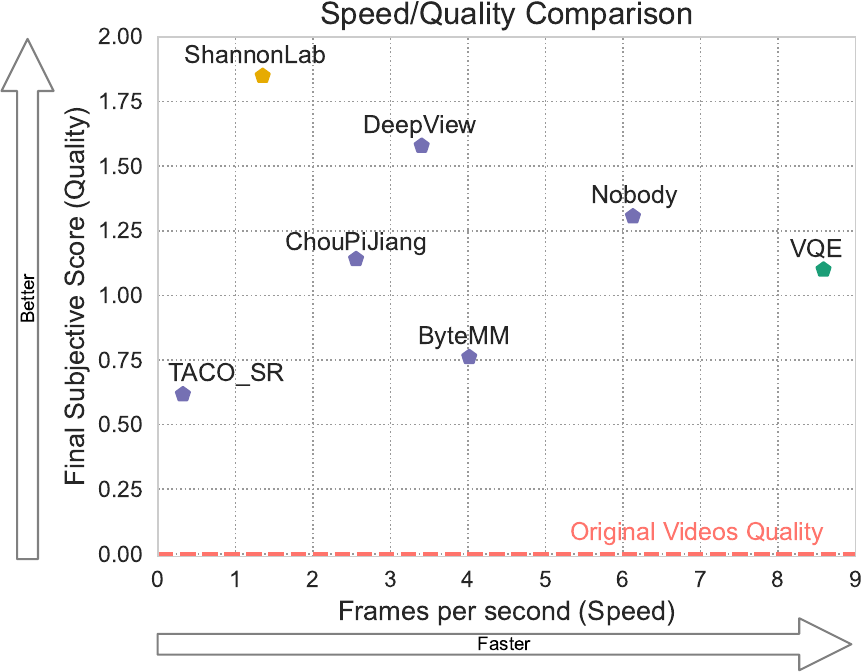}
    \caption{Visualization of the final leaderboard subjective scores and measured algorithm speeds. The most efficient solution is highlighted in green, while the team with the highest-quality solution is marked in orange.}
    \label{fig:speed}
\end{figure}

In recent years, UGC (User-Generated Content) videos have become widespread due to the popularity of short-form video platforms like Kwai, TikTok, and others. However, since these videos are typically captured by non-professionals, they often suffer from lower subjective quality, with issues such as unstable footage, poor lighting, and compression artifacts. Delivering the most visually pleasing content has become an important task for the video platforms, as it greatly impacts viewer interest. Developing UGC video enhancement benchmarks might be very helpful and drive the rapid advancement of video processing techniques for UGC.

UGC videos often suffer from distortions such as motion blur, noise, faded colors, low resolution, and compression artifacts. Therefore, many video enhancement methods have been proposed~\cite{chan2022basicvsr++, Haris2020, Liu2018, Liu2020}. In addition, for the specific image and video enhancement tasks in recent years, various algorithms have been developed to address these issues. Methods for reducing motion blur have been proposed in~\cite{li2023self, liu2024motion, Kupyn_2018_CVPR, chen2022simple}. Color correction and contrast enhancement have been explored in~\cite{Getreuer2012ACE, Gupta2016AGCCPF, Wang2005BPHEME, Lv2018MBLLEN, Zheng2022SGZSL, Zhang2022DCCNet}, in the past few years, there has been plenty of work on enhancing the quality of compressed video \cite{yang2017decoder, yang2018enhancing, wang2017novel, lu2018deep, yang2020learning, guan2019mfqe, deng2020spatio, huo2021recurrent, wang2020multi, xu2019non, yang2019quality}. Additionally, some other works address different types of distortions. To address these tasks, several datasets, challenges and benchmarks have been introduced in recent years~\cite{gao2023vdpve, agarla2023quality, yang2021ntire, yang2021ntire2, nah2021ntire, voronin2024new}.

This challenge aims to provide a platform for researchers and industry professionals to develop and evaluate algorithms for enhancing UGC videos, focusing on subjective quality assessment. The objectives of this UGC Video Enhancement Challenge are to establish a benchmark for UGC video enhancement, including real-world videos with diverse distortions and encourage the development of algorithms that improve the perceptual quality of UGC videos, ensuring better viewing experiences across various content types and capture conditions. 

In this competition, we propose a dataset of 150 videos to evaluate the submitted methods. An additional 40 videos were given to participants at the beginning of the competition to familiarize themselves with the content, but were not used in the evaluation. Some videos in the dataset were collected from the real users of short-form video platform, while others were specifically recorded by users based on predefined scenarios. This dataset includes diverse content types and capture conditions, reflecting the real-world challenges of UGC video enhancement. 

Challenge dataset was split into training, validation, and testing sets with sizes of 40, 20, 20, 20, and 90 videos for training, three validation and test stages, respectively. From the test set, 60 videos were available to the participants, the remaining 30 were in a private subset and were released only after the end of the competition. Methods results on these 30 videos were obtained by the organizers team independently by running the code of the participants solutions in the final phase of the competition. During the competition, participants has access only to subjective assessments results, but did not see the enhanced videos of other teams. For the final test phase, validation videos were also included in the evaluation process. Thus, the final evaluation dataset consists of 150 videos. 

To evaluate participants methods at all phases of the competition, we largely relied on the \href{https://subjectify.us/}{Subjectify.us} platform. For evaluation, we used pairwise subjective comparisons and aggregated the scores using the Bradley-Terry model \cite{bradley1952rank}. This approach ensures a robust subjective assessment of video enhancement methods by leveraging both direct quality comparisons and ranking-based score estimation.

The competition consists of three development stages and a final test stage, attracting a total of 79 registered participants. Across these stages, 26 teams participated. Ultimately, 7 teams provided fact sheets and passed source code verification in the final stage. Descriptions of the methods proposed by the participants are provided in \cref{teams_and_methods}.


This challenge is one of the NTIRE 2025~\footnote{\url{https://www.cvlai.net/ntire/2025/}} Workshop associated challenges on: ambient lighting normalization~\cite{ntire2025ambient}, reflection removal in the wild~\cite{ntire2025reflection}, shadow removal~\cite{ntire2025shadow}, event-based image deblurring~\cite{ntire2025event}, image denoising~\cite{ntire2025denoising}, XGC quality assessment~\cite{ntire2025xgc}, night photography rendering~\cite{ntire2025night}, image super-resolution (x4)~\cite{ntire2025srx4}, real-world face restoration~\cite{ntire2025face}, efficient super-resolution~\cite{ntire2025esr}, HR depth estimation~\cite{ntire2025hrdepth}, efficient burst HDR and restoration~\cite{ntire2025ebhdr}, cross-domain few-shot object detection~\cite{ntire2025cross}, short-form UGC video quality assessment and enhancement~\cite{ntire2025shortugc,ntire2025shortugc_data}, text to image generation model quality assessment~\cite{ntire2025text}, day and night raindrop removal for dual-focused images~\cite{ntire2025day}, video quality assessment for video conferencing~\cite{ntire2025vqe}, low light image enhancement~\cite{ntire2025lowlight}, light field super-resolution~\cite{ntire2025lightfield}, restore any image model (RAIM) in the wild~\cite{ntire2025raim}, raw restoration and super-resolution~\cite{ntire2025raw} and raw reconstruction from RGB on smartphones~\cite{ntire2025rawrgb}.

\begin{table*}[t]
\tabcolsep=8pt
\centering
\caption{Challenge leaderboard: subjective ranking based on Bradley-Terry scores. Zero scores corresponds to the original (without enhancement) video, confidence intervals computed relative to original video. FPS are based on the organizers' measurements.}
\begin{tabular}{c|l|c|c|c|c}
\toprule
Rank & \multicolumn{1}{c|}{Team}        & Final Score ± (95\% CI) & Public Score ± (95\% CI) & Private Score ± (95\% CI) & FPS   \\
\midrule
1    & ShannonLab  & 1.848 ± 0.060     & 1.780 ± 0.066            & 2.149 ± 0.141             & 1.039 \\
2    & DeepView    & 1.578 ± 0.058     & 1.482 ± 0.064            & 1.998 ± 0.139             & 2.617 \\
3    & Nobody      & 1.305 ± 0.057     & 1.273 ± 0.064            & 1.452 ± 0.134             & 4.714 \\
4    & ChouPiJiang & 1.140 ± 0.057     & 1.087 ± 0.063            & 1.371 ± 0.133             & 1.966 \\
5    & VQE         & 1.100 ± 0.057     & 1.043 ± 0.063            & 1.345 ± 0.133             & 6.605 \\
6    & ByteMM      & 0.761 ± 0.056     & 0.716 ± 0.063            & 0.960 ± 0.132             & 3.089 \\
7    & TACO\_SR    & 0.617 ± 0.057     & 0.618 ± 0.063            & 0.618 ± 0.132             & 0.247 \\ \bottomrule
\end{tabular}
\label{tab:subjective_scores}
\end{table*}

\section{Challenge}
\label{sec:formatting}
The NTIRE 2025 UGC Video Enhancement Challenge is organized to drive advancements in video enhancement techniques for user-generated content. This challenge focuses on improving the perceptual quality of UGC videos through novel restoration and enhancement methods. By establishing a new benchmark, the challenge aims to guide future research and development in this field. The following sections provide details on the challenge, including dataset, evaluation protocols, and competition phases.

\subsection{Dataset}

The new dataset was provided to ensure a reliable and comprehensive evaluation of each method. For this purpose, we collected two subsets: (1) videos obtained from a short-form UGC video platform and (2) videos recorded by users of Yandex Tasks (a crowdsourcing platform) following predefined scenarios. Users were asked to record indoor or outdoor scenes under various lighting conditions, capturing subjects such as animals, people, vehicles, portraits, and food. To ensure diversity in the evaluation dataset, the combined set was divided into 20 clusters based on precomputed VQMT measures for blurring, noise, brightness flickering, blocking, and spatial and temporal information. From these clusters, a demonstration training set of 40 videos, three validation sets of 20 videos each, and a testing set of 90 videos were manually selected, including 30 private sequences, which were not available to the participants.

\subsection{Evaluation}

Subjective comparison was used for the method ranking. The evaluation process included three validation stages during the contest and a final subjective test to determine the winners. Subjective votes were collected using crowdsourcing for both the validation and the final evaluation. For the evaluation, we used side-by-side preference selection, with the following instruction for participants: 

"\textit{You will be shown pairs of videos with different quality. You need to select in each pair the video with the most acceptable quality for viewing, or note that in this pair the quality is almost the same}". 

Thus, participants could select from three options: "left", "right" or "can't choose". Each participant completed 20 pairs, 2 of which were validation ones with predefined answers. Validation questions were obtained by compressing two original videos from a dataset with a high crf value. The pairs were assigned randomly to each assessor, the participants did not know which questions were validation ones, and also did not know how the videos were obtained. Only votes from performers who passed both verification questions were selected. Then the matrix of pairwise votes was randomly balanced so that each pair had exactly 10 votes. In total, we collected votes from over 8000 crowdsourcing participants.

To obtain rank scores from pairwise votes, we used the Bradley-Terry model \cite{bradley1952rank}, which assumes that the probability of video \( i \) being preferred over video \( j \) is given by:  

\[
P(i \succ j) = \frac{e^{s_i}}{e^{s_i} + e^{s_j}}
\]

where \( s_i \) and \( s_j \) are the desired subjective score estimates of videos \( i \) and \( j \), which are derived by maximizing the likelihood of the observed pairwise comparisons. 

The final ranking of the solutions was determined based on the estimated scores $\hat{\boldsymbol{s}}_i$ in Table~\ref{tab:subjective_scores}, providing a fair and statistically grounded evaluation of all submissions.

Additionally, we compute $95 \%$ confidence intervals for score differences $s_i-s_j$, using the asymptotic normality of the maximum likelihood estimates (MLE). Specifically, if $\hat{s}_i$ and $\hat{s}_j$ are the MLEs of the scores, then their difference is approximately normal with variance estimated from the inverse Fisher information matrix $I_Y^{-1}(\hat{\theta})$, where $\hat{\theta} = \{\hat{s_1}, \hat{s_2}, \hat{s_3},...\}$. The standard error of $\hat{s}_i-\hat{s}_j$ computed as: 
$$\hat{s}_{i j}=\sqrt{\left(I_Y^{-1}(\hat{\theta})\right)_{i i}+\left(I_Y^{-1}(\hat{\theta})\right)_{j j}-2\left(I_Y^{-1}(\hat{\theta})\right)_{i j}}
$$
\noindent and the resulting 95\% confidence interval is given by:
$\hat{s}_i-\hat{s}_j \pm z_{score}(0.025) \hat{s}_{i j}$
\noindent where $z_{score}(0.025) \approx 1.96$ is the critical value from the standard normal distribution. In Table~\ref{tab:subjective_scores} we provide 95\% CI relative to the original video (without enhancement), i.e. $\hat{s}_{Original}$, which was also involved in all pairwise comparisons.

During each validation stage, 20 new validation videos were provided. Three validation sets allowed participants to evaluate different solutions and receive feedback on their quality. The final evaluation dataset consisted of 60 videos matching the validation set, 60 test videos provided to participants, and 30 hidden test videos. For the hidden dataset, videos were generated using the participants' code. Only open and reproducible results were considered.

As the task was to create methods that not only improve the perceptual quality of UGC videos but also ensure that the enhanced results retain high visual quality after being recompressed using x265 at 3000 kbps (standard bitrate value for transmitting by short-form video platforms), making the challenge both practical and impactful. We transcoded videos with FFmpeg using following command: \texttt{ffmpeg -i input\_path -c:v libx265 -preset fast -b:v 3000k -pix\_fmt yuv420p -an output\_path}\newline

All submitted solutions were tested on the same hardware with the following specifications:
\begin{itemize}
    \item CPU: 2 $\times$ Intel Xeon Silver 4216 CPU @ 2.10GHz
    \item RAM: 188 GB
    \item GPU: NVIDIA TITAN RTX
\end{itemize}
\section{Results}

\begin{figure}[ht]
    \centering
    \includegraphics[width=0.99\columnwidth]{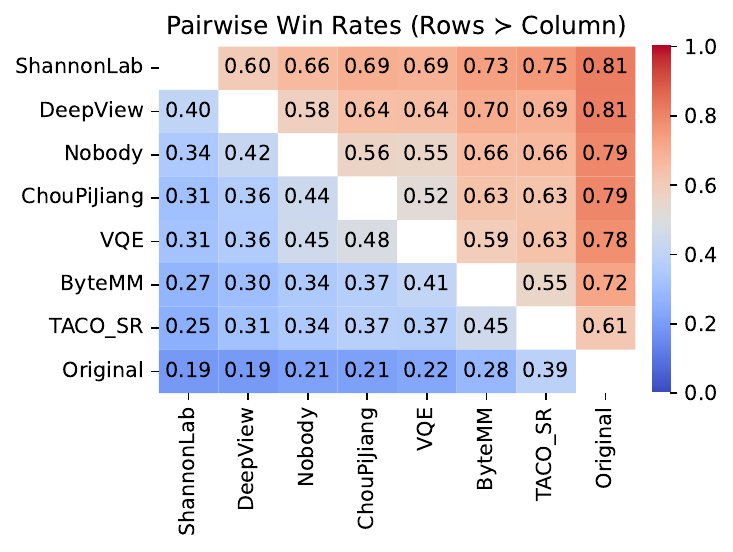}
    \caption{Pairwise comparison matrix with winning rates of participants' methods, rows over columns.}
    \label{fig:winrate}
\end{figure}

In the first validation phase, 17 participants submitted results. The second validation phase received 19 submissions, and the third had 20 submissions. The results of the validation phases are presented in the challenge repository. For the final scoring, we received 7 valid submissions. A summary of the methods used by the participating teams is presented in Section~\ref{sec:meth}, while team details are provided in Section~\ref{sec:aff}.

Table~\ref{tab:subjective_scores} presents the scores and rankings for the final submissions of participated teams in the subjective comparisons. The table includes rankings for the overall evaluation (150 videos) as well as separately for the public (120 videos) and private (30 videos) dataset parts, demonstrating consistency between dataset segments. Additionally, Fig.~\ref{fig:speed} illustrates the overall results, including solution speed, while Fig.~\ref{fig:winrate} provides a preference matrix showing the fraction of times each method was preferred in pairwise comparisons.

\section{Teams and Methods}\label{teams_and_methods}
\label{sec:meth}
\subsection{ShannonLab}
\subsubsection{Framework}

\begin{figure}[h]
    \centering
    \includegraphics[width=1.0\columnwidth]{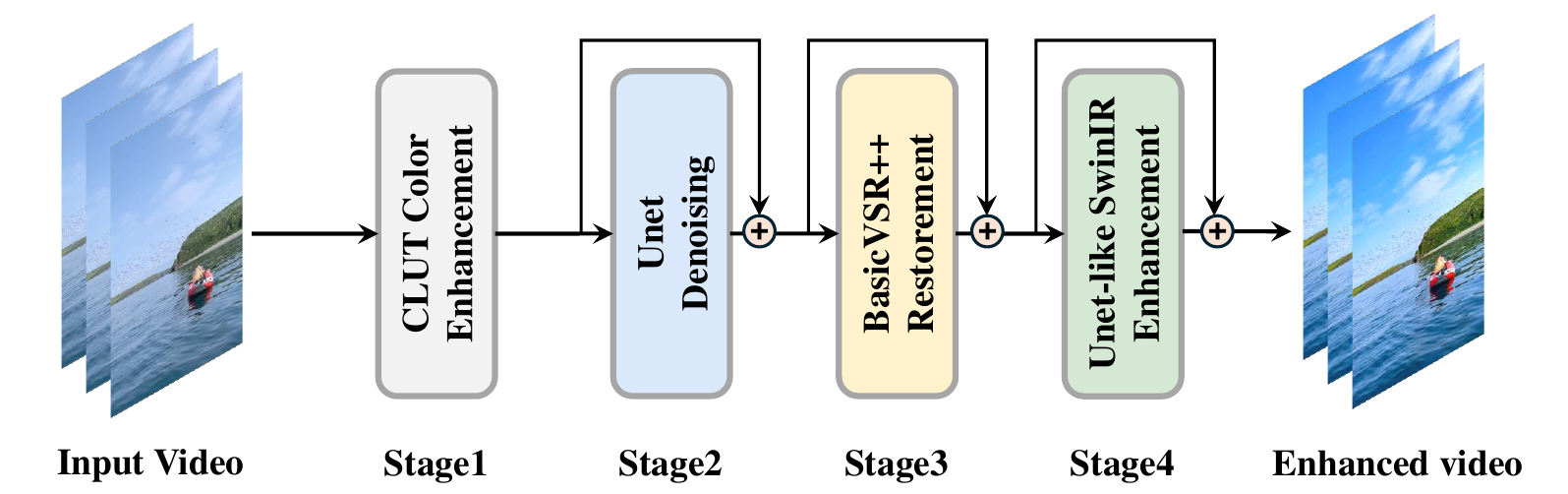}
    \caption{ShannonLab solution: Progressive training of a multi-Stage framework for UGC video restoration.}
    \label{fig:ShannonLab}
\end{figure}

We propose a multi-stage progressive training framework for UGC video restoration (TRestore), as shown in Fig.~\ref{fig:ShannonLab}. 

The main idea of our framework is that enhancing UGC videos is a complex task, which we decompose into four stages for progressive processing. 

In the first stage, we focus on enhancing colors. To achieve this, we apply  CLUT \cite{zhang2022clut}, which enhances color through adaptive prediction of LUT. Compared to other methods, this approach performs better in inference speed and robustness.
In the second stage, our goal is to remove noise, especially compression artifacts and ISP noise, which may have a bad impact on video encoding. Therefore, we use a lightweight U-Net network to remove the noise.
In the third stage, the network is developed on top of BasicVSR++ \cite{chan2022basicvsr++}, which stabilizes the results in the temporal domain to make sure good performance even when compressed to 3000kbps. 
In the last stage, we further improve the quality of the enhanced consecutive frames by a image restoration network, i.e., SwinIR \cite{liang2021swinir}. Additionally, we modified the structure of SwinIR to be similar to U-Net, allowing us to achieve faster inference speeds with the same number of parameters. This stage helps mitigate severe blurry and further improve quality.

Finally, the four stages are cascaded to produce the final results. Besides, to make the network and to prevent degradation caused by an excessive number of stages, we apply residual connections between stages 2, 3, and 4.

\subsubsection{Training}
To train the four-stage network, we used a large number of public datasets such as LDV3 \cite{yang2022aim} and REDS \cite{liu2022video}. Besides, for UGC videos, we download an amount of 4K videos from Pexels \cite{pexels2025}. To simulate actual degradation methods, we modeled camera sensor noise, color degradation (saturation and contrast), compression artifacts, motion blur and scaling operations, and we randomized various degradation methods when creating the degraded data. 
There are more training details for each stage:

\textbf{Stage1:} Only the CLUT is trained using L1 loss for 600k iterations. learning rate was set to 1e-4, with a batch size of 32 and a patch size of 720.

\textbf{Stage2:} Only the denoising U-net is trained using L2 loss for 600k iterations. learning rate was set to 1e-4, with a batch size of 32 and a patch size of 640.

\textbf{Stage3:} Only BasicVSR++ is trained for 120k iterations using: $$L2 + 1 \ast PerceptualLoss + 0.1 \ast GANLoss$$ The learning rate was set to 2e-4, with batch size 8, patch size 512, and the number of frames is 30.

\textbf{Stage4:} BasicVSR++ and Unet-like SwinIR are trained  for 120k iterations using: $$L2 + 0.1 \ast PerceptualLoss + 0.01 \ast GANLoss + 4 \ast LPIPS$$. The learning rate was set to 1e-5, with batch size 8, patch size 512, and the number of frames is 30.

\subsubsection{Inference}
During inference, we implemented two optimization strategies to improve objective evaluation metrics.

\textbf{Color Enhancement:} 
To achieve better subjective effects, we amplified the color residuals obtained by CLUT. The corresponding coefficient is 1.2, which obtaining more vivid results.

\textbf{Feature Interpolate:} 
We perform inference with a segment of 30 frames. However, jitter often occurs between segments. To solve it , we interpolate the features before the upsampling layer of BasicVSR++ between two segments and then restore them into images.


\subsection{DeepView}

\subsubsection{Framework}
\begin{figure}[h]
    \centering
    \includegraphics[width=1.0\columnwidth]{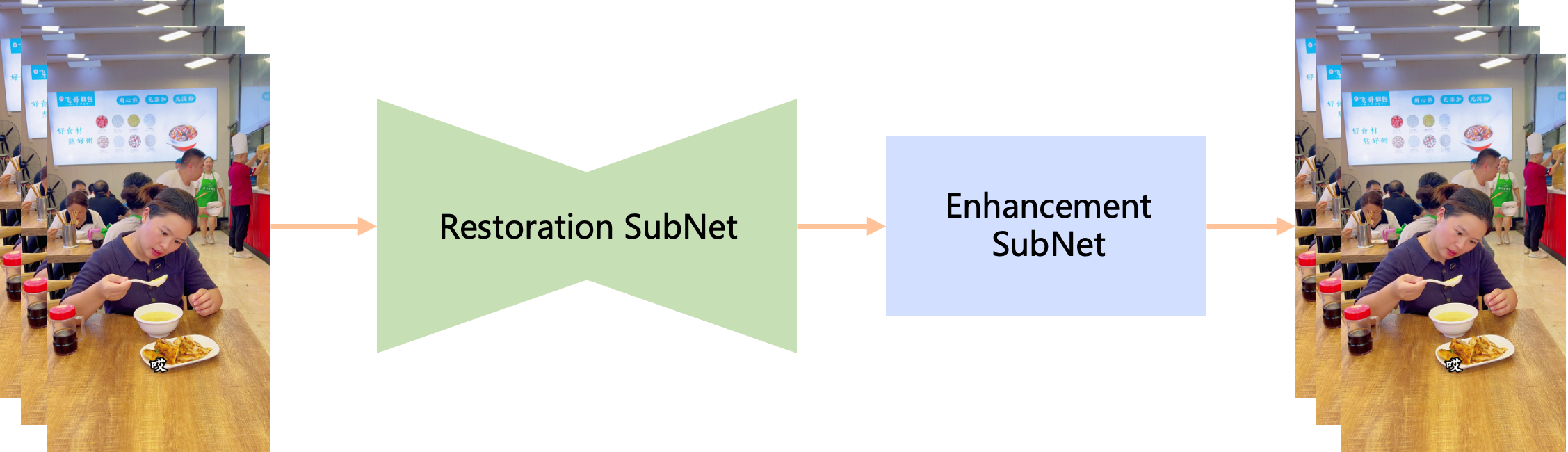}
    \caption{DeepView solution: Two stage UGC video restoration framework.}
    \label{fig:frameworkDeepview}
\end{figure}
We proposed the video enhancement framework to address the inherent challenges of User-Generated Content (UGC) videos, including noise, compression artifacts, and visual inconsistencies. The methodology is structured into two cascaded stages: degradation restoration and texture refinement. This dual-stage approach ensures a balance between computational efficiency and perceptual quality, delivering visually appealing results while minimizing processing overhead.

The first stage focuses on restoring the low-level distortions commonly present in UGC videos. These distortions include noise, compression artifacts, uneven illumination, and color shifts that degrade the visual quality of the content. To address these issues, we employ a lightweight U-Net architecture with skip connections, specifically designed for efficient and robust restoration. The network extracts features at multiple scales. This allows the network to simultaneously address both local artifacts (e.g., blocky compression noise) and global degradations (e.g., color casts or uneven lighting). Skip connections between the encoder and decoder ensure that fine-grained details are preserved during the restoration process. What's more, the contracting path of the network is equipped with spatial attention mechanisms that effectively suppress mixed noise sources, such as sensor noise and encoding artifacts. This ensures that the restored video is free from distracting visual noise while retaining important structural details. The first stage has 26 convolutional layers that can expand receptive fields and perform global adjustments to brightness and color consistency. This capability allows the network to correct color shifts and uneven illumination, restoring natural and visually consistent tones across the video.

The second stage focuses on generative enhancement, aiming to recover high-frequency details and realistic textures that are often lost during video capture or compression. This stage is implemented using a deep network composed of 15 cascaded residual blocks, each enhanced with dense connections and channel attention modules. The stacked ResBlocks progressively refine the video features, with attention mechanisms prioritizing semantically important regions, such as facial features, textures, and fine details. This ensures that the enhanced video exhibits realistic and visually appealing textures.

\subsubsection{Training}

To train our two-stage network, we used a combination of public datasets, including LDV3 \cite{yang2022aim}, REDS \cite{liu2022video}. These datasets provided a diverse range of video content, ensuring that our model was exposed to various types of distortions and artifacts commonly found in UGC videos. For realistic degradation simulation, we modeled mixed distortions to create training data that closely resembled real-world scenarios. The training data for the first stage incorporated color distortions, such as random saturation shifts and contrast adjustments. For the second stage, the training data included randomized degradations such as Poisson-Gaussian noise, motion blur, and H.265/H.264 compression. The degradation parameters were dynamically sampled per batch to improve robustness. In stage 1, the U-Net was trained with a hybrid loss function combining L1 loss and Perceptual Loss to balance pixel accuracy and semantic consistency, over 600,000 iterations with a batch size of 32 and a patch size of 512$\times$512. The initial learning rate was set to $1\mathrm{e}{-4}$ and halved every 10,000 iterations, using the Adam optimizer with $\beta_1=0.9$ and $\beta_2=0.99$. 

For the second stage, the sub-network was optimized using a combination of L2 loss, perceptual loss, LPIPS and GAN loss to enhance textures without over-smoothing, over 300,000 iterations with a batch size of 16 and a patch size of 512$\times$512. After pretraining both stages independently, we jointly fine-tuned the network for an additional 50,000 iterations with reduced learning rates, e.g. $1\mathrm{e}{-5}$. To prevent overfitting, we applied spatial augmentations such as rotation, flipping, and chromatic aberration, as well as temporal jittering techniques like frame dropping and shuffling. By following these detailed training protocols and incorporating diverse data sources and realistic degradation simulations, our two-stage network was robustly trained to enhance UGC videos effectively, ensuring high perceptual quality and computational efficiency.

\subsubsection{Test}

We evaluated 120 videos on the Tesla A10 GPU, including the time required for video reading, writing, and preprocessing, which amounted to a total of 13,272.26 seconds. The videos comprised a total of 21,825 frames, resulting in an average processing speed of 1.8 frames per second (fps). When considering only the model's inference speed, the processing rate for 720p videos was 5.73 fps, while for 1080p videos, it was 2.5 fps.

\subsection{Nobody}

\subsubsection{Framework}
\begin{figure}[h]
    \centering
    \includegraphics[width=1.0\columnwidth]{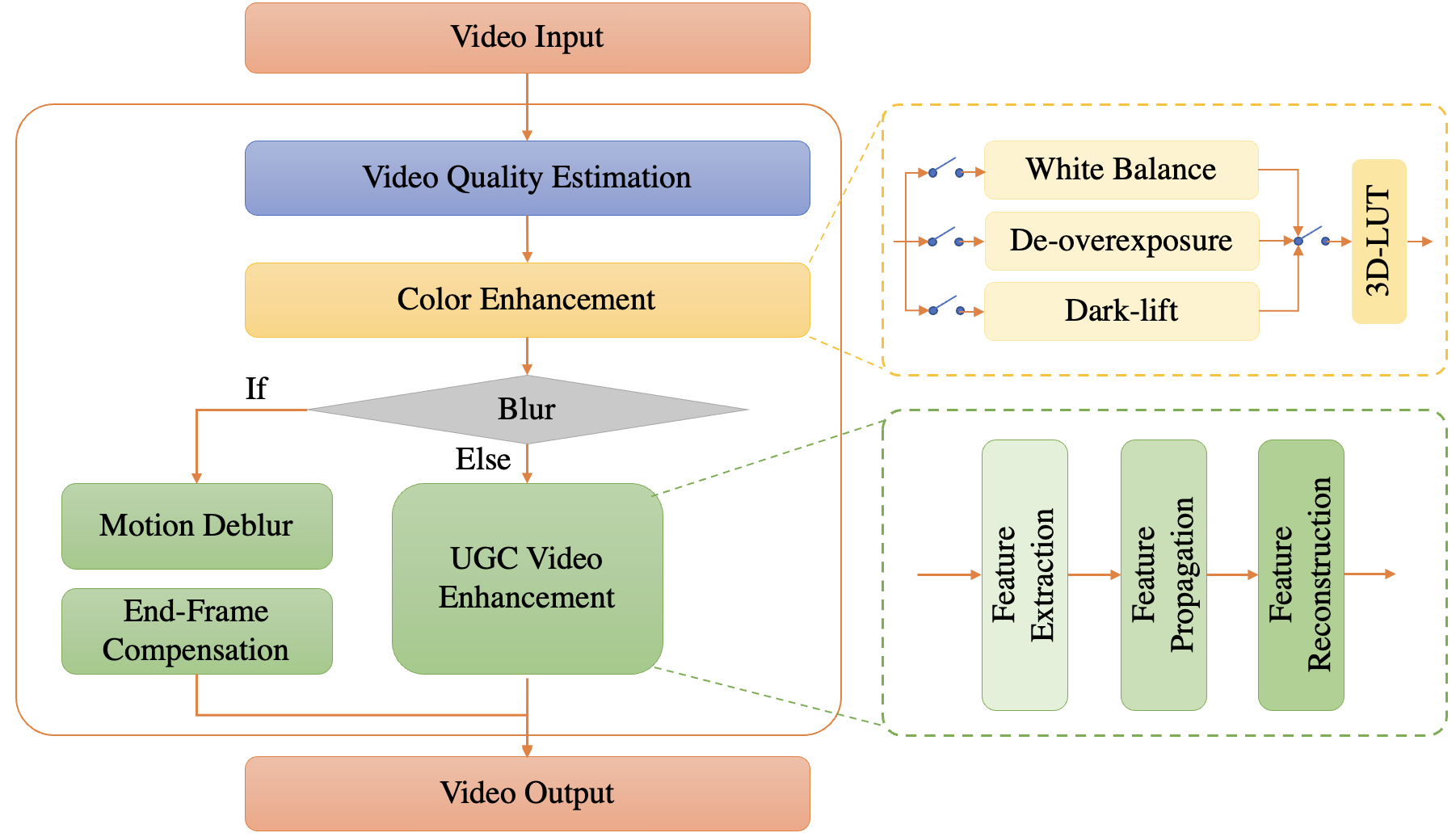}
    \caption{Nobody solution: video enhancement pipeline.}
    \label{fig:piplineNobody}
\end{figure}

Observing that the UGC videos quality are different, besides severe compression artifacts, may also contain dark light, de-focus blur, motion blur, and noise. We propose a two - stage approach for the UGC video enhancement, and use multiple operators in each stage. The first stage for color enhancement, and the second stage for artifacts compression as well as de-noise, de-blur and texture enhancement. Our main pipeline is shown in Fig.~\ref{fig:piplineNobody} \\We estimate the video quality and distortion type first, and process the video according to the estimated type. In the first stage, we use 3D-LUT \cite{zeng2020lut} and some maching learning methods, and the second stage are two GAN models based on Real-ESRGAN \cite{wang2021realesrgan} framework, additionally we observed that UGC videos shot with handheld devices often shake at the last frame, so we add end-frame compenstation by flow \cite{DBLP:journals/corr/RanjanB16}.

\subsubsection{Details}
In the first stage, we used CNN as well as machine learning algorithms such as white balance and gamma correction. For the CNN model, we use Adobe 5k \cite{fivek} and self-made pictures about 10k as training data. The self - made data are generated by diffusion with prompts such as "High-contrast", "bright color" and degeneration them as LQ when trainging. \\
In the second stage, we use FFHQ \cite{karras2019stylegan} and about 1000 high-quality videos from YouTube, we degenerate the high-quality videos by ffmpeg, with parameter -crf from 24 to 36, and multiple blur kernels. In addition, we use online degradation methods twice as \cite{wang2021realesrgan} when training. For the face in FFHQ, we randomly paste them in the training image pairs. The training Loss is:  $$L_{total} = L_1 + 0.1 \times L_{LPIPS} + 0.05 \times L_{GAN}$$ with a learning rate of $2 \times 10^{-4}$.

\subsection{ChouPiJiang}
\subsubsection{Framework}
Our solution is based on Real-ESRGAN \cite{wang2021realesrgan} and the network is RRDBNet. We Use 70,000 FFHQ \cite{karras2019stylegan} in the wild and 200 4K YouTube videos as training data.

\subsubsection{Test}
\begin{figure}[h]
    \centering
    \includegraphics[width=1.0\columnwidth]{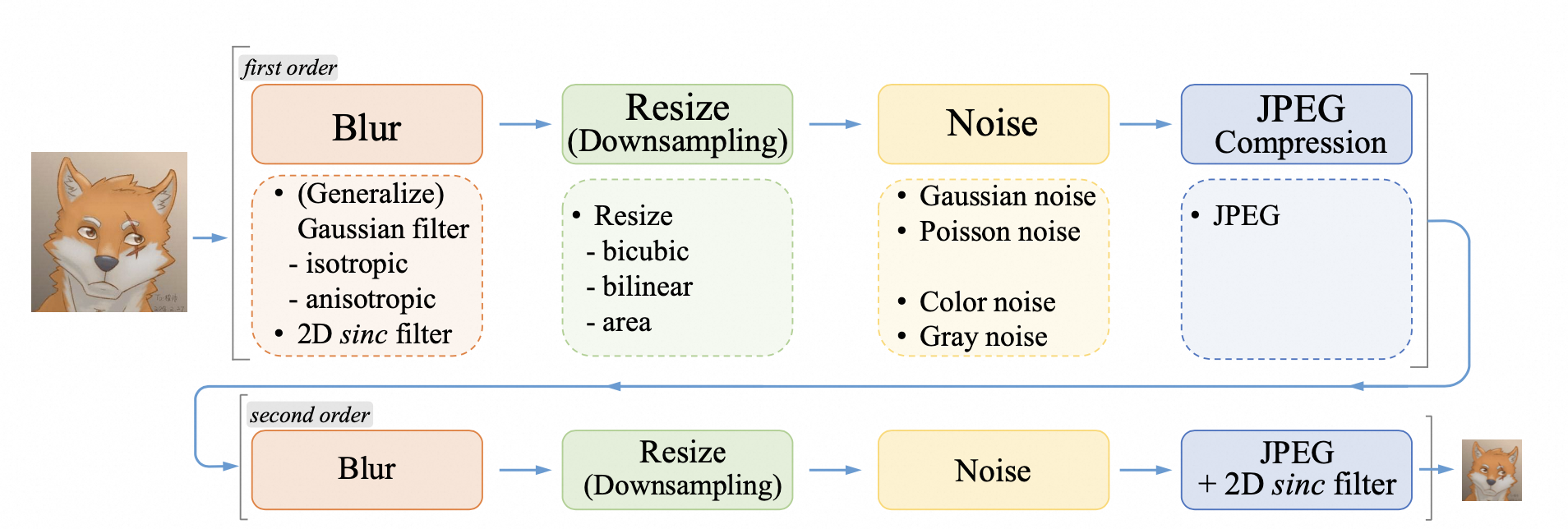}
    \caption{ChouPiJiang solution: The degradation pipeline for network training.}
    \label{fig:pipelineChouPiJiang}
\end{figure}

We use a second-order degradation process
to model more practical degradations same as Real-
ESRGAN \cite{wang2021realesrgan}, The pipeline of the second-order degradation
process is shown in Fig.~\ref{fig:pipelineChouPiJiang}.

\subsubsection{Network Detail}
\begin{figure}[h]
    \centering
    \includegraphics[width=1.0\columnwidth]{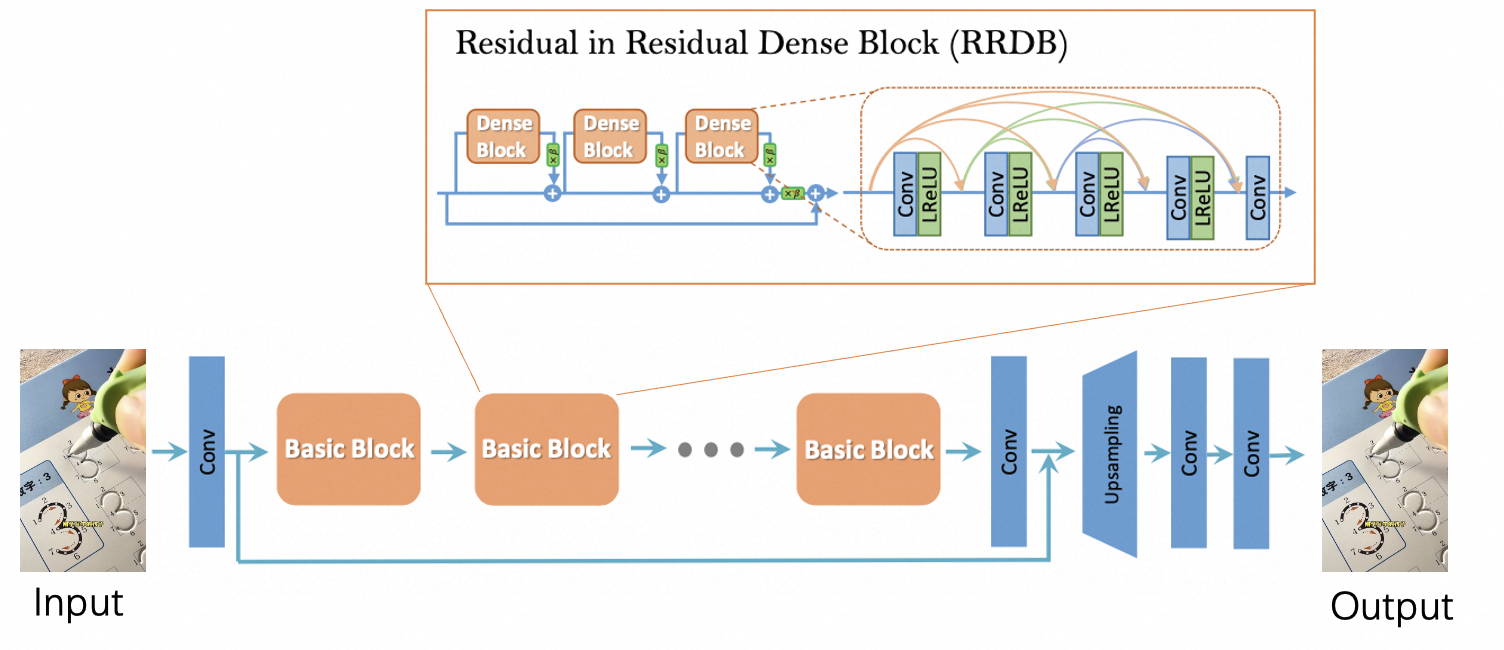}
    \caption{ChouPiJiang solution: the network architecture.}
    \label{fig:networkChouPiJiang}
\end{figure}

Our Network based on RRDBNet from ESRGAN \cite{wang2018esrgan} as shown in Fig.~\ref{fig:networkChouPiJiang}. with channel = 128, growth = 32, and blocks = 23.

\subsection{VQE}
\begin{figure}[h]
    \centering
    \includegraphics[width=1.0\columnwidth]{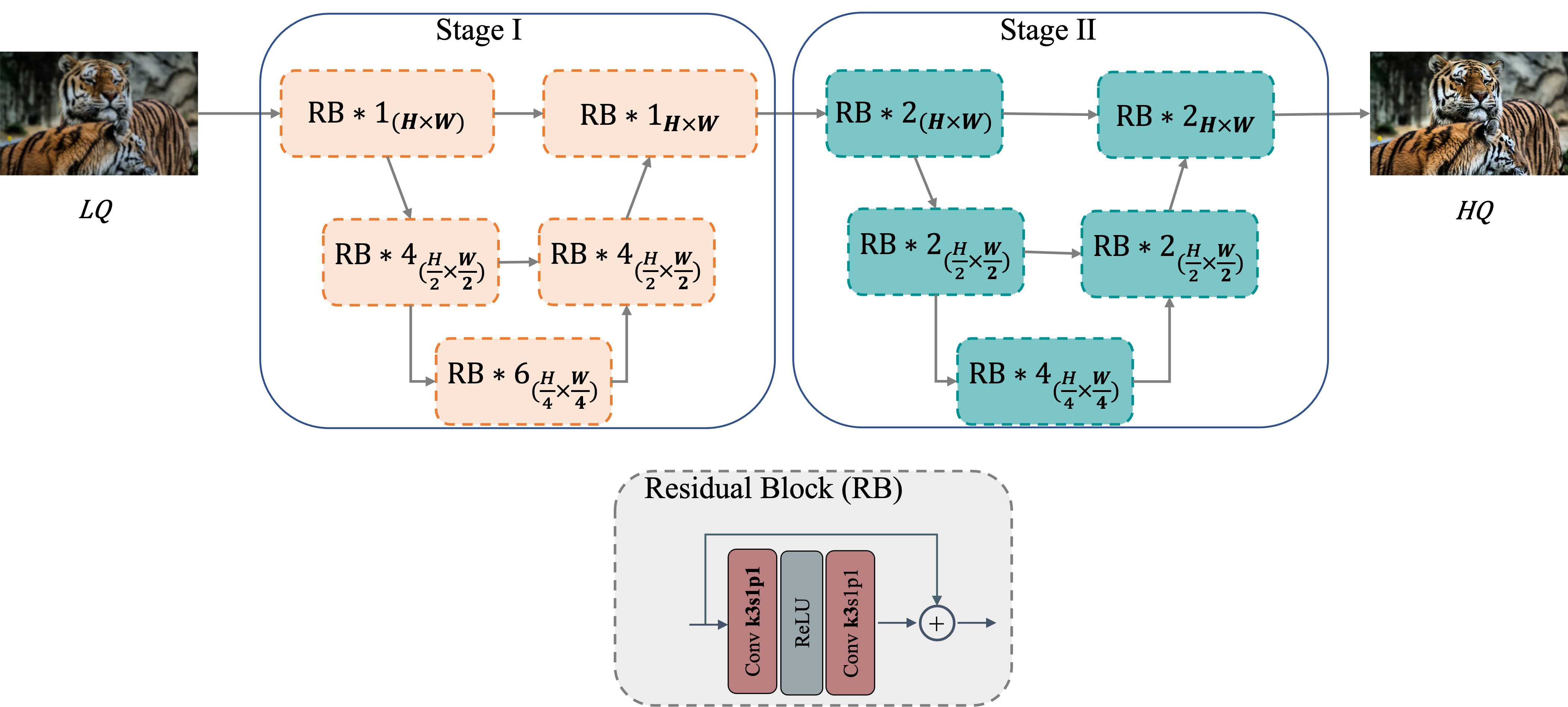}
    \caption{VQE solution: the network architecture.}
    \label{fig:frameworkVQE}
\end{figure}

As shown in \ref{fig:frameworkVQE}, our algorithm employs a two-stage processing approach. In the first stage, referred to as Model 1, the primary focus is on removing severe degradations present in the video.  The second stage, referred to as Model 2, is designed to effectively enhance the sharpness and clarity of the video.

\subsection{ByteMM}

\begin{figure}[h]
    \centering
    \includegraphics[width=1.0\columnwidth]{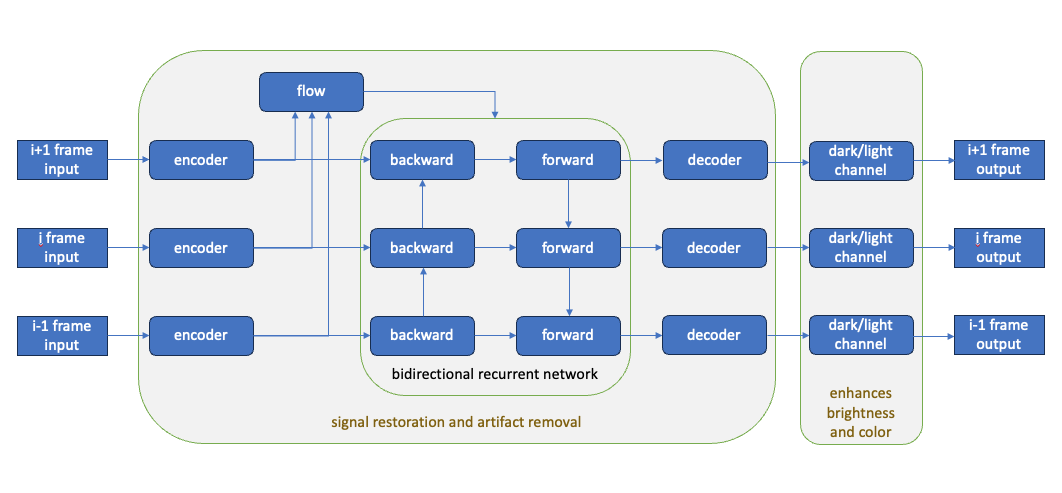}
    \caption{ByteMM solution: pipeline scheme.}
    \label{fig:bytemm}
\end{figure}

Our UGC video enhancement approach consists of two stages. An illustration is provided in Fig.~\ref{fig:bytemm}.

In the first stage, we employ a modified version of RealBasicVSR for signal restoration and artifact removal. The primary structural differences from the original RealBasicVSR lie in its more lightweight and integrated design. During training, various artificial degradation synthesis methods are used to generate high-quality (HQ) and low-quality (LQ) video pairs. This is the only stage that requires training. Specifically, we optimize the model for common UGC content such as faces and text. To address the spatially non-uniform degradation caused by user post-processing (e.g., subtitles and effects), we incorporate targeted design in the artificial degradation process, making the model more suitable for UGC video enhancement.

The second stage enhances brightness and color based on dark channel and bright channel priors. A non-deep-learning method is adopted to improve stability and robustness. Since this stage has very few hyperparameters, all of which have clear physical meanings, it does not require training and only needs manual tuning. In the color enhancement process, we specifically restrict adjustments to skin tones to preserve the original semantic integrity of UGC videos.

\subsection{TACO\_SR}
Inspired by recent advances in image generation using diffusion models~\cite{rombach2022high}, diffusion-based approaches~\cite{wu2024one} have achieved significant progress in the field of image restoration. We propose two stage PiNAFusionNet for UGC video enhancement. 

\subsubsection{Network Architecture.}  
\begin{figure}[h]
    \centering
    \includegraphics[width=1.0\columnwidth]{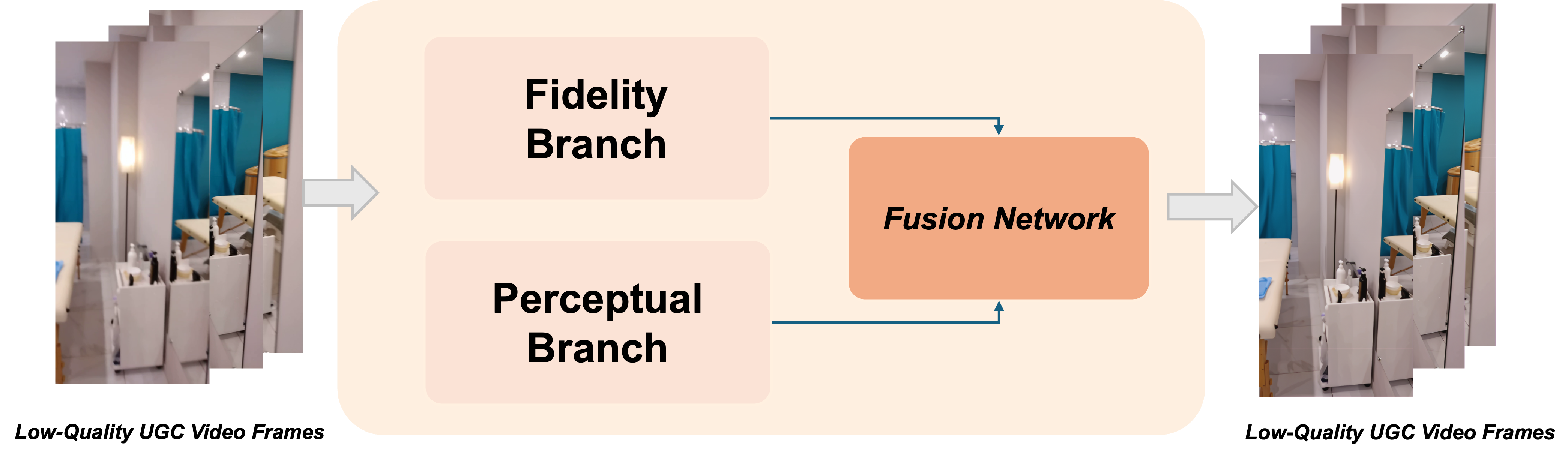}
    \caption{TACO\_SR solution: network architecture.}
    \label{fig:TACO_SR}
\end{figure}

The overall architecture of PiNAFusion-Net is illustrated in Fig.~\ref{fig:TACO_SR}. The model is composed of two stages. In the first stage, we employ a dual-branch structure consisting of a \textit{Fidelity Branch} and a \textit{Perceptual Branch}, both based on an adjustable super-resolution network that emphasizes either pixel-level or semantic-level perception. These branches produce two complementary outputs, which are subsequently processed by a \textit{Fusion Network}. In second stage, a filter is used to extract fine-grained details from the first stage result, forming an initial enhanced representation. The Fusion Network employs initial enhanced representation and a trainable module to produce the detail-enhanced frame.

\subsubsection{Training Details.} The proposed model is implemented in PyTorch and optimized using the AdamW optimizer with an initial learning rate of $1 \times 10^{-5}$. Given the scarcity of high-quality paired video datasets for UGC video enhancement, we resort to training on synthetic paired short-form UGC images.

\section{Teams and Affiliations}\label{affilations}
\label{sec:aff}
\noindent \textit{\textbf{Team:}}\\
Organizers \\ 
\textit{\textbf{Members:}}\\
Nikolay Safonov$^{1,2}$ (nikolay.safonov@graphics.cs.msu.ru), Alexey Bryncev$^1$, Andrey Moskalenko$^{1,2,3}$, Dmitry Kulikov$^{1,2}$, Dmitry Vatolin$^{1,2,4}$, Radu Timofte$^5$\\
\textit{\textbf{Affiliations:}}\\
$^1$: Lomonosov Moscow State University, Russia\\
$^2$: MSU Institute for Artificial Intelligence, Russia\\
$^3$: AIRI, Moscow, Russia\\
$^4$: Innopolis University, Russia\\
$^5$: Computer Vision Lab, University of Würzburg, Germany\\

\noindent \textit{\textbf{Team:}}\\
ShannonLab\\
\textit{\textbf{Members:}}\\
Haibo Lei$^1$ (hypolei@tencent.com), Qifan Gao$^1$, Qing Luo$^1$, Yaqing Li$^1$.   \\
\textit{\textbf{Affiliations:}}\\
$^1$:Tencent, China\\


\noindent \textit{\textbf{Team:}}\\
DeepView\\
\textit{\textbf{Members:}}\\
Jie Song (724215288@qq.com), Shaozhe Hao.   \\


\noindent \textit{\textbf{Team:}}\\
NoBody\\
\textit{\textbf{Members:}}\\
Meisong Zheng$^1$ (1377855931@qq.com), Jingyi Xu$^{1,2}$ , Chengbin Wu$^{1}$, Jiahui Liu$^{1}$, Ying Chen$^{1}$, Xin Deng$^{2}$, Mai Xu$^{2}$\\
\textit{\textbf{Affiliations:}}\\
$^1$: Department of Tao Technology, Alibaba Group, China\\
$^2$: Beihang University, Beijing, China  \\


\noindent \textit{\textbf{Team:}}\\
ChouPiJiang\\
\textit{\textbf{Members:}}\\
Peipei Liang$^1$ (jjjin1990@gmail.com), Jie Ma$^2$, Junjie Jin$^3$\\
\textit{\textbf{Affiliations:}}\\
$^1$: Longyuan (Beijing) New Energy Engineering Technology Co., Ltd., China\\
$^2$: China Telecom Digital Intelligence Technology Co., Ltd., China\\
$^3$: Key Laboratory of Optical Astronomy, National Astronomical Observatories, Chinese Academy of Sciences, China\\

\noindent \textit{\textbf{Team:}}\\
VQE\\
\textit{\textbf{Members:}}\\
Yingxue Pang$^1$ (pangyx@mail.ustc.edu.cn) \\
\textit{\textbf{Affiliations:}}\\
$^1$: University of Science and Technology of China, China\\


\noindent \textit{\textbf{Team:}}\\
ByteMM\\
\textit{\textbf{Members:}}\\
Fangzhou Luo (luofangzhou@bytedance.com), Yingxue Pang, Kai Chen, Shijie Zhao\\
\textit{\textbf{Affiliations:}}\\
MMLab, ByteDance Inc\\

\noindent \textit{\textbf{Team:}}\\
TACO\_SR\\
\textit{\textbf{Members:}}\\
Mingyang Wu$^1$ (mingyang@tamu.edu), 
Renjie Li$^1$, 
Yushen Zuo$^{1,2}$, 
Shengyun Zhong$^3$, 
Zhengzhong Tu$^1$\\
\textit{\textbf{Affiliations:}}\\
$^1$: Texas A\&M University, USA\\
$^2$: The Hong Kong Polytechnic University, Hong Kong\\
$^3$: Northeastern University, USA\\

\section*{Acknowledgments}
This work was partially supported by the Humboldt Foundation. We thank the NTIRE 2025 sponsors: ByteDance, Meituan, Kuaishou, and University of Wurzburg (Computer Vision Lab).

The evaluations for this research were carried out using the MSU-270 supercomputer of Lomonosov Moscow State University.
{
    \small
    \bibliographystyle{ieeenat_fullname}
    \bibliography{main}
}


\end{document}